\documentclass[letterpaper, 10 pt, conference]{ieeeconf}
\IEEEoverridecommandlockouts
\usepackage{cite}
\usepackage{amsmath,amssymb,amsfonts}
\usepackage{algorithm}
\usepackage{algpseudocode}
\usepackage{graphicx}
\usepackage{textcomp}
\usepackage[dvipsnames]{xcolor}
\usepackage{caption}
\usepackage{subcaption}
\usepackage{verbatim}
\usepackage{relsize}
\usepackage{array}
\usepackage{tikz}
\usetikzlibrary{positioning, decorations.pathreplacing, shapes}

\def\BibTeX{{\rm B\kern-.05em{\sc i\kern-.025em b}\kern-.08em
    T\kern-.1667em\lower.7ex\hbox{E}\kern-.125emX}}
    
\renewcommand{\vec}[1]{
\textbf{\textit{#1}}}

\newcommand\overeq[1]{\mathrel{\overset{\makebox[0pt]{\mbox{\normalfont\tiny\sffamily \emph{#1}}}}{=}}}

\title{\LARGE \bf
IoT Federated Blockchain Learning at the Edge
}

\author{James Calo$^{1}$ and Benny Lo$^{2}$
\thanks{$^{1}$James Calo is with The Hamyln Centre / Department of Computing,
        Imperial College London, London, UK
        {\tt\small jam414@ic.ac.uk}}%
\thanks{$^{2}$Benny Lo is with The Hamyln Centre / Department of Surgery and Cancer, Imperial College London,
        London, UK
        {\tt\small benny.lo@imperial.ac.uk}}%
}

\begin{document}

\maketitle
\thispagestyle{empty}
\pagestyle{empty}

\suppressfloats
%
%

\begin{abstract}

IoT devices are sorely underutilised in the medical field, especially within machine learning for medicine, yet they offer unrivalled benefits. IoT devices are low cost, energy efficient, small and intelligent devices \cite{RN82}.
\par
In this paper, we propose a distributed federated learning framework for IoT devices, more specifically for IoMT (Internet of Medical Things), using blockchain to allow for a decentralised scheme improving privacy and efficiency over a centralised system; this allows us to move from the cloud based architectures, that are prevalent, to the edge.
\par
The system is designed for three paradigms: 1) Training neural networks on IoT devices to allow for collaborative training of a shared model whilst decoupling the learning from the dataset \cite{RN80} to ensure privacy \cite{RN78}. Training is performed in an online manner simultaneously amongst all participants, allowing for training of actual data that may not have been present in a dataset collected in the traditional way and dynamically adapt the system whilst it is being trained.
2) Training of an IoMT system in a fully private manner such as to mitigate the issue with confidentiality of medical data and to build robust, and potentially bespoke \cite{RN70}, models where not much, if any, data exists.
3) Distribution of the actual network training, something federated learning itself does not do, to allow hospitals, for example, to utilize their spare computing resources to train network models.

\end{abstract}


\section{Introduction}

The healthcare industry could save up to \$300 billion by focusing more on IoMT devices, especially when dealing with chronic illnesses \cite{RN81}; therefore, we can expect to see IoMT as common place in the healthcare industry proving an undeniable incentive to utilise these ubiquitous and wide spread devices in a distributed, secure and intelligent manner.
\par
To this end, distributed machine learning on the edge is a persuasive solution that leverages the trend technology in the healthcare industry is following; unfortunately, current systems fail to consolidate these abilities and instead focus on distinct aspects.
In this paper, we propose a novel approach to address the challenge of training machine learning systems, in particular neural networks (NNs), on the devices themselves, which we refer to as learning on the edge (LotE).
Our system provides enhanced, if not total, privacy and security, which given the sensitive nature of the patient data is essential, in a distributed and robust manner by default.
This is achieved by combining federated learning, a paradigm that aggregates individually trained networks, and blockchain to remove the need for a centralised server.
\par
\begin{figure}[t]
\begin{subfigure}{0.3\columnwidth}
 \includegraphics[width=\textwidth]{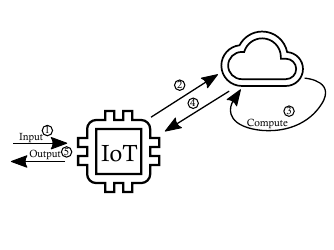}
    \caption{Cloud \\ (Remote Server)}
    \label{fig:TCA}
\end{subfigure}
\begin{subfigure}{0.3\columnwidth}
 \includegraphics[width=\textwidth]{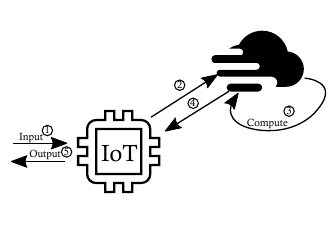}
    \caption{Fog \\ (Local Server)}
    \label{fig:FA}
\end{subfigure}
\begin{subfigure}{0.3\columnwidth}
 \includegraphics[width=\textwidth]{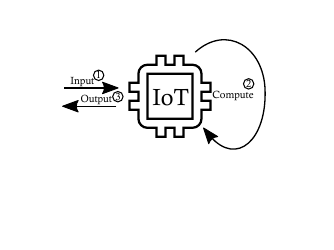}
\caption{On The Edge \\ (No Server)}
\label{fig:OTEA}
\end{subfigure}
\caption{Comparison of Cloud, Fog and Edge Architectures}
\label{fig:cloud_paradigms}
\end{figure}

There is a clear hierarchy of architectural archetypes; on one end there is cloud computing 
\cite{
RN18, 
RN19, 
RN6
}
containing vast resources with increased computational power. This, however, requires communication between the local system and the cloud.
Therefore issues such as loss of connection, network congestion, cyber security, etc. will affect the system's performance. 
On the other end is edge computing 
\cite{
RN25,  
RN95   
}
with restricted resources but unparalleled access to the device as both the system and the backing computations are on the same (or physically close) device. In between these archetypes is fog computing 
\cite{
RN95,
RN14}; having the same structure as cloud computing but instead using a local server as shown in Fig. \ref{fig:cloud_paradigms}.
Current solutions target either cloud and fog based computing or do in fact run on the edge; however, they either lack machine learning or only perform inferencing, and not training, which is the most computational demanding task.
\par

\suppressfloats

One ideal use of IoMT devices is mobile health (mHealth). In developing countries, this has been shown as an effective method to monitor patients; unfortunately, these systems are often unintelligent relying on basic mobile phone functionality \cite{RN6}.
These approaches purposefully avoid internet connection, since it is unreliable in many developing countries, yet the required components are in place to leverage the power of machine learning.
\par
On the other hand, the focus of mHealth in developed countries is for smart wearable devices, often paired with an app, yet these too are unintelligent and use a fraction of the ability of modern IoT systems running their computations via the cloud; this allows less capable hardware to run complex computations but suffers from latency issues and must be connected to the internet to work which is not ideal.
\par
Whilst there exists a handful of IoT systems that aim to leverage machine learning on the edge, they only support inferencing and not training.


For example, the STM32CubeAI converts neural networks to run on STM32 Arm Cortex-M-based microcontrollers \cite{RN17} and has been used to create a human activity recognition (HAR) fitness tracker embedding a convolutional neural network (CNN) in a wrist worn, low power, MCU for inferencing \cite{RN8}. Frameworks such as these are a step in the right direction but suffer from the need to train the models on a dedicated system or the cloud.
\par
Simultaneously, there have been advances in the hardware required to infer, and potentially train, on the edge. GPUs are better adapted to machine learning methods than CPUs but are rarely found in embedded devices and not all GPUs were created equal; the majority of the frameworks utilise the CUDA language which is designed specifically for Nvidia GPUs. Furthermore, GPUs may be usurped by AI accelerator application-specific integrated circuits such as Google's TPU (Tensor Processing Unit) and FPGAs (field-programmable gate array), which are used in Microsoft's Project Brainwave to improve real-time deep neural network (DNN) inferencing \cite{RN18, RN19}. The requirement for specific hardware increases the physical size, power draw and cost of devices; this is counterproductive for IoMT where smaller and less obtrusive devices are preferred. By moving the learning to the edge on a CPU one can upgrade existing devices whilst keeping the footprint of newer devices smaller and focus more on efficiency.
\par
The infrastructure required to take IoMT and edge/fog computing to the next level is already in place in a hospital. The users only move within a set area and data collection happens in the same location meaning federated learning is ideally suited to edge learning in a hospital \cite{RN25}; multiple surgeries happen simultaneously and can all learn together to train models to increase generalizability, improving the model's overall performance by treating each patient or surgery as a decentralised dataset whilst still allowing for bespoke training on a per patient basis \cite{RN70}. This is ideally suited to clinical settings as federated learning never shares data thereby keeping data private and allowing training on previously inaccessible tasks such as that of anastomotic leak detection where the existing data, of which there is little, is severely unbalanced.


\section{Methods}

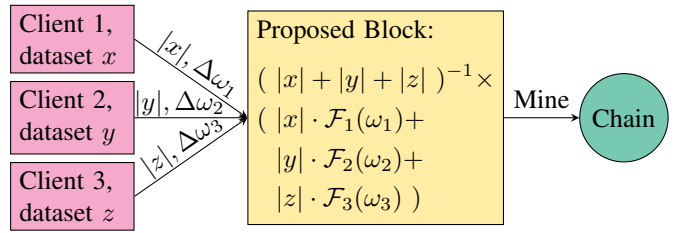
\begin{figure}[t]
\begin{tikzpicture}[
networkUpdate/.style={draw, text width=4em, fill=Rhodamine!50, minimum size=6mm},
]
 
\node[networkUpdate] (nu1) at (-0.2,1){Client 1, dataset $x$};
\node[networkUpdate, below=1mm of nu1] (nu2) {Client 2, dataset $y$};
\node[networkUpdate, below=1mm of nu2] (nu3) {Client 3, dataset $z$};
 
\node [draw,
    fill=Goldenrod!50,
    minimum width=25mm,
    minimum height=1.2cm,
    right=15mm of nu2,
    text width=9em
]  (NewBlock) 
{
Proposed Block:
\begin{align*} 
( \ &|x|+|y|+|z| \ )^{-1} \times \\
( \ &|x| \cdot  \mathcal{F}_1(\omega_1) + \\ 
    &|y| \cdot  \mathcal{F}_2(\omega_2) + \\
    &|z| \cdot  \mathcal{F}_3(\omega_3) \ )
\end{align*}
};
 
\node [draw,
    circle,
    fill=SeaGreen!75, 
    right=10mm of NewBlock
] (chain) {Chain};
 
\path[-stealth] 
(nu1.east) edge node[sloped, anchor=center, above=-1mm]{$|x|, \Delta \omega_1$} (NewBlock.west)

    
(nu3.east) edge node[sloped, anchor=center, above=-1mm] {$|z|, \Delta \omega_3$} (NewBlock.west);
   
\draw[-stealth] (nu2.east) -- (NewBlock.west)
   node[above=-1.5mm]at (1.225,0){$|y|, \Delta \omega_2$};

 
\draw[-stealth] (NewBlock.east) -- (chain.west) 
    node[midway,above]{Mine};

\end{tikzpicture}

\caption{Example of three clients contributing to the blockchain using (\ref{global_training}). Note $( \ x \cup y \cup z \ ) \subseteq all\_possible\_data$. }
\label{fig:SystemArch}
\end{figure}

In order to address the issues discussed, we combine federated learning with blockchain to enable the computing resources in a hospital environment to train neural networks whilst ensuring security and privacy (Fig. \ref{fig:SystemArch}). Therefore, the main contributions of this work are:
\begin{enumerate}
    \item We propose a blockchain framework for use with machine learning models, either directly through our framework or via the C API which allows users to use other popular machine learning frameworks, such as TensorFlow.
    
    \item We develop a novel federated learning system, for training at the edge (LotE), that is fully decentralised, leveraging our blockchain framework, ensuring that the data is private and secured against malicious attacks and requiring no trust between participants.
    
    \item Using our federated learning system, we develop a configurable system for the training of neural networks on IoT devices, trained online, requiring only a small percentage to be active at any one time; this enables devices that run infrequently or on a schedule to still participate without hindering the training process.
    
    \item We propose a novel method for distributing the training step of each individual device (prior to the federated step) to another device (node) without losing the privacy guarantee of federated learning; a reconstructable form of the training data never leaves the device. This would allow hospitals to harvest spare computing power, e.g. from a receptionist's PC.
\end{enumerate}

\subsection{Federated learning}

The goal for every model in the system, both the local models and the federated (global) model is to minimize the loss with respect to the model parameter:

\vspace{-1.25em}

\begin{equation}
\label{minimise_loss}
\min_{\omega \in \mathbb{R}^d} \mathcal{F}(\omega) =
\ell(\vec{x}, \vec{y}, \omega)
\end{equation}

\vspace{-0.5em}

Where $\ell$ is a chosen loss function, consistent across all participating models, $\vec{x},\vec{y}$ are the training (input and desired output) vectors and $\omega$ is the model's parameters.
\par
However, for the federated model, which has seen no training data (insuring the data privacy), these local updates are then collated, in our case via a block to be added to the blockchain, and averaged based on the \textit{FedAvg} algorithm \cite{RN27}. 
Hence, for each local model index $m = 1,2,....,M$ of a participating IoT device, which has performed local training (\ref{minimise_loss}) either on the device itself or via a processing node, over its training set $\mathcal{T}_m$:

\vspace{-1.75em}

\begin{equation}
\label{global_training}
\mathcal{F}_{federated}(\omega) \triangleq \frac{1}{|\mathcal{T}|}
\sum_{m=1}^{M}|\mathcal{T}_m| * \mathcal{F}_m(\omega_m)
\end{equation}

\vspace{-0.5em}

Where $\mathcal{T} = \sum_{m=1}^{M}|\mathcal{T}_m|$. Whilst this (\ref{global_training}) can be simplified mathematically, practically we need to scale the contribution of the local update to the global by how many examples it has seen ($\mathcal{T}_m$).

\subsection{Decentralisation with blockchain}

One of the greatest weaknesses of vanilla federated learning is the requirement on a centralised server; to address this we propose to use blockchain to tweak the paradigm to a decentralised distributed ledger. This additionally shifts the logical architecture from the cloud/fog, which essentially comprises of devices connected to a server, to the edge, where every device is independent and autonomous; the system will work even with only one node and even if all nodes go down, the system can recover fully since each node contains a copy of the accepted blockchain, this may not be possible if the central server lost its data. The following details our design regarding the fundamental components of the blockchain.

\subsubsection{Block}
Our block format closely mirrors bitcoin's format but with two major changes: The target formula and the federated components. The target is used to decide when the block has been mined. 
We use proof of work (PoW) over alternative, more green (both environmentally and chronologically) consensus mechanisms, such as proof of stake (PoS) 
\cite{RN104,
RN106}; the downside would be the energy cost. A miner who performs PoW must continually guess, in a deterministic manner, a hash that is less than the target, since the target is in big endian hexadecimal format we refer to it as a hash with more leading zeros than the target.
\par
The criticism stems from the high computational cost that provide no actual benefit, other than to make it infeasible for a malicious node to pervert the system. However, in an IoT system this is actually beneficial over PoS; for example, with hundreds of mining devices, the problem can be split across them, much like mining pools. 
Moreover, since the blockchain is being utilized as a trust mechanism for federated learning, the mining target difficulty can remain lower, reducing the computational cost and increasing the rate at which blocks are added to the chain; this results in lower powered devices having enough computing resources to generate hashes competitively whilst still providing the same protection. We therefore decided on adding a block approximately every 1.5 minutes; this is long enough for multiple local updates, from different sources, to be added to the block, prior to the block being added to the chain, without being so long that either the global update is outdated or a local device that misses the update will grow stale. PoS would not be as suitable since it relies too heavily on transactions, doesn't include mining, would give too much power to larger institutions and it promotes coin hording which negates the bonus benefit of blockchain, rewards: This is what incentivizes hospitals to utilise their spare computing power.  

\subsubsection{Mining}
In order to mine local updates via PoW we have to store the target in the block; however, the true target size is the same as the hash and so, much like bitcoin, we encode the target in 4 bytes:

\vspace{-2em}

\begin{equation}
\label{target}
\vec{Target} \overeq{hex}
0\textrm{x}\overbrace{\phi_1\phi_2}^{\mathlarger{\Phi}} \overbrace{\theta_1\theta_2\theta_3\theta_4\theta_5\theta_6}^{\mathlarger{\Theta}} \triangleq \Theta * 2^{8 * (\mathlarger{\Phi} - 4)}
%
\end{equation}
Such that the first byte ($\phi_1\phi_2$) is an exponential scale and the lower three bytes contribute the linear scale. As with bitcoin we use 8 to scale the exponential, as there are 8 bits in a byte which simplifies a lot of the bit manipulations; however, we use an exponential scale value of 4 (as opposed to bitcoin's 3) in order to generate targets with more usable values at the lower range.

\subsubsection{Cryptography}
We currently use the same cryptographic hashes as bitcoin, SHA256 for mining and RIPEMD160 for transactions, and use double hashing.

\subsubsection{Networking}
The peer-to-peer (P2P) network is possibly the most vital component; whilst a single node can still be functional, the benefits of federated learning would be severely reduced. The system needed to handle two cases, 1) obtaining a copy of the blockchain and a list of addresses of other nodes, to which the node will send their own address and 2) broadcasting information to other nodes. By using a pair of UDP sockets, we not only parallelize the communication we can split the two cases across different devices; for example, a hospital may have many IoMT devices but none with networking capabilities, just Bluetooth; they could therefore connect all IoMT devices to a single or pair of network enabled IoT devices which would handle the networking and correct forwarding, much like Network Address Translation (NAT) with regards to WiFi routers. Consequently, any IoT device can participate as long as they can connect to a networking node, e.g. via Bluetooth, hardwired to a communication module etc, somewhere down the line. Additionally, if one set of devices are all training on the same set of data, only one device needs to connect to the outbound UDP connection and, as long as everyone connects to the inbound connection, all devices gain the benefits.

\section{Results}

To test our system\footnotemark{} we used TensorFlow to build a simple model, comprising two convolutional and max pooling layers, a final convolutional layer and two dense layers, to classify the CIFAR-10 dataset. Using both standard and federated training paradigms we trained the model five times using 10\%, 25\%, 50\%, 75\% and 100\% of the training data; in the federated case the data was shared equally amongst all participating models, such that no two models saw the same data points, as would be the case in a live system (especially when using image data as the input). For each subset of the data, the model was trained for 150 epochs whilst additionally measuring the affect of altering the number of epochs each participating member trained for before the federated update and the number of participates to the federated scheme as shown in Table \ref{table:fedVsStd}. The resulting accuracies in each sub-table are all within a small range showing that federated learning produces similar results to the standard method but with the benefit of being applicable to distribution and working on different (albeit similar) datasets with no shared datapoints. Furthermore, when training using federated updates, with each ```local training''' round being performed sequentially, the training process runs quicker compared with the standard method due to each participant operating on smaller subsets of the data, allowing for optimisations such as better caching; this is especially apparent on smaller devices.

\footnotetext{We will release the code used to generate these results in a python notebook on GitHub once this paper is accepted for publication.}

\begin{table}[t]
    \begin{subtable}[h]{0.48\columnwidth}
        \centering
    \begin{tabular}{|c|c|c|} 
        \hline
        Update & \#Models & Accuracy \\
        \hline
        25  & 2 & 50.57\% \\
        25  & 4 & 51.04\% \\
        25  & 8 & 51.33\% \\
        \hline
        50  & 2 & 50.37\% \\
        50  & 4 & 51.3\% \\
        50  & 8 & 52.33\% \\
        \hline
        75  & 2 & 50.95\% \\
        75  & 4 & 49.03\% \\
        75  & 8 & * 52.90\% * \\
        \hline
        \multicolumn{2}{|c|}{Non-Federated}
        & 52.04\% \\
        \hline
    \end{tabular}
     \caption{Trained with 10\% of the training data}
        \label{tab:data10}
     \end{subtable}
     \hfill
     \begin{subtable}[h]{0.48\columnwidth}
        \centering
    \begin{tabular}{|c|c|c|} 
        \hline
        Update & \#Models & Accuracy \\
        \hline
        25  & 2 & 57.69\% \\
        25  & 4 & 57.01\% \\
        25  & 8 & 57.87\% \\
        \hline
        50  & 2 & 57.81\% \\
        50  & 4 & 57.95\% \\
        50  & 8 & 58.16\% \\
        \hline
        75  & 2 & 58.54\% \\
        75  & 4 & * 59.51\% * \\
        75  & 8 & 59.43\% \\
        \hline
        \multicolumn{2}{|c|}{Non-Federated}
        & 56.94\% \\
        \hline
    \end{tabular}
     \caption{Trained with 25\% of the training data}
        \label{tab:data25}
     \end{subtable}
     \\
     \begin{subtable}[h]{0.48\columnwidth}
        \centering
     \begin{tabular}{|c | c | c|} 
        \hline
        Update & \#Models & Accuracy \\
        \hline
        25  & 2 & 63.16\% \\
        25  & 4 & * 64.29\% * \\
        25  & 8 & 63.53\% \\
        \hline
        50  & 2 & 63.06\% \\
        50  & 4 & 64.24\% \\
        50  & 8 & 62.33\% \\
        \hline
        75  & 2 & 63.93\% \\
        75  & 4 & 63.64\% \\
        75  & 8 & 63.89\% \\
        \hline
        \multicolumn{2}{|c|}{Non-Federated}
        & 62.62\% \\
        \hline
    \end{tabular}
     \caption{Trained with 50\% of the training data}
        \label{tab:data50}
     \end{subtable}
     \begin{subtable}[h]{0.48\columnwidth}
        \centering
     \begin{tabular}{|c | c | c|} 
        \hline
        Update & \#Models & Accuracy \\ 
        \hline
        25  & 2 & 65.91\% \\
        25  & 4 & 66.06\% \\
        25  & 8 & 65.83\% \\
        \hline
        50  & 2 & 66.46\% \\
        50  & 4 & 65.65\% \\
        50  & 8 & 65.76\% \\
        \hline
        75  & 2 & 65.61\% \\
        75  & 4 & * 66.86\% * \\
        75  & 8 & 65.51\% \\
        \hline
        \multicolumn{2}{|c|}{Non-Federated}
        & 65.88\% \\
        \hline
    \end{tabular}
     \caption{Trained with 75\% of the training data}
        \label{tab:data75}
     \end{subtable}
     \begin{subtable}[h]{\columnwidth}
        \centering
     \begin{tabular}{|c | c | c|} 
        \hline
        Update & \#Models & Accuracy\\
        \hline
        25  & 2 & 68.09\% \\
        25  & 4 & 68.75\% \\
        25  & 8 & 68.85\% \\
        \hline
        50  & 2 & * 69.23\% * \\
        50  & 4 & 68.12\% \\
        50  & 8 & 68.11\% \\
        \hline
        75  & 2 & 68.56\% \\
        75  & 4 & 68.09\% \\
        75  & 8 & 68.07\% \\
        \hline
        \multicolumn{2}{|c|}{Non-Federated}
        & 67.64\% \\
        \hline
    \end{tabular}
     \caption{Trained with 100\% of the training data}
        \label{tab:data100}
     \end{subtable}
    \caption{Accuracy of convolutional model on CIFAR-10 after 150 epochs. The top accuracy in each case is highlighted.}
    \label{table:fedVsStd}
\end{table}

\section{Discussion}
A new distributed learning approach is proposed with the aim of allowing learning on the edge by designing a light weight, distributed, autonomous system that is a natural fit for IoT devices that are abundant, particularly in a hospital environment. We have produced a fully functional system that allows for training of neural networks either through our APIs or popular machine learning frameworks, such as TensorFlow allowing for existing networks to become federated. These can then be run across a multitude of IoT devices to build a universal and general model in an extremely secure and privacy enhancing manner. These are vital requirements in a clinical situation where datasets are difficult to come by, are severely limited in size and may not be shareable.
\par
However, there are a few components that we would like to address in future work: In order to allow spare computing resources to be shared to distribute the ```local training''' there needs to be a way to secure the training data either by working on an encrypted form (homomorphic encryption) or by converting the data into a non-reversible representation, for example Fourier or wavelet transformation. Furthermore, smart contract would be invaluable for automating tasks and sharing of processing capabilities.

\bibliographystyle{IEEEtran}
\bibliography{IEEE_EMBC_2023}

\end{document}